\documentclass[journal]{IEEEtran}

\ifCLASSOPTIONcompsoc
  \usepackage[nocompress]{cite}
\else
  \usepackage{cite}
\fi
\ifCLASSINFOpdf
\else
\fi
\usepackage{graphicx}
\usepackage{amsmath}
\usepackage{amssymb}
\usepackage{booktabs}
\usepackage{bbding}
\usepackage{color,xcolor}
\usepackage{multirow}
\usepackage{float}
\usepackage{adjustbox}
\usepackage{url}
\usepackage{diagbox}
\usepackage[switch]{lineno}
\usepackage[hidelinks]{hyperref}
\usepackage[capitalize]{cleveref}
\usepackage{soul}

\newcommand{\modi}[1]{\textcolor{black}{#1}}
\newcommand{\add}[1]{\textcolor{black}{#1}}
\newcommand{\chen}[1]{\textcolor{black}{#1}}
\newcommand{\yq}[1]{\textcolor{black}{#1}}

\begin{document}

\newcommand{\reflabel}{dummy} % Dummy initial reflabel - use renewcommand...

% \newcommand{\be}{\begin{equation}}
% \newcommand{\ee}{\end{equation}}
% \newcommand{\eqlabel}[1]{\label{eq:\reflabel-#1}}
% \renewcommand{\eqref}[2][\reflabel]{(\ref{eq:#1-#2})}

% Generic reference commands
\newcommand{\seclabel}[1]{\label{sec:\reflabel-#1}}
\newcommand{\secref}[2][\reflabel]{Section~\ref{sec:#1-#2}}
\newcommand{\Secref}[2][\reflabel]{Section~\ref{sec:#1-#2}}
\newcommand{\secrefs}[3][\reflabel]{Sections~\ref{sec:#1-#2} and~\ref{sec:#1-#3}}

\newcommand{\eqlabel}[1]{\label{eq:\reflabel-#1}}
\renewcommand{\eqref}[2][\reflabel]{(\ref{eq:#1-#2})}
\newcommand{\Eqref}[2][\reflabel]{(\ref{eq:#1-#2})}
\newcommand{\eqrefs}[3][\reflabel]{(\ref{eq:#1-#2}) and~(\ref{eq:#1-#3})}

\newcommand{\figlabel}[2][\reflabel]{\label{fig:#1-#2}}
\newcommand{\figref}[2][\reflabel]{Fig.~\ref{fig:#1-#2}}
\newcommand{\Figref}[2][\reflabel]{Fig.~\ref{fig:#1-#2}}
\newcommand{\figsref}[3][\reflabel]{Figs.~\ref{fig:#1-#2} and~\ref{fig:#1-#3}}
\newcommand{\Figsref}[3][\reflabel]{Figs.~\ref{fig:#1-#2} and~\ref{fig:#1-#3}}

\newcommand{\tablelabel}[2][\reflabel]{\label{table:#1-#2}}
\newcommand{\tableref}[2][\reflabel]{Table~\ref{table:#1-#2}}
\newcommand{\Tableref}[2][\reflabel]{Table~\ref{table:#1-#2}}
\newcommand{\etal}{\emph{et al.}}
\newcommand{\eg}{e.g.}
\newcommand{\ie}{i.e. }
\newcommand{\etc}{etc. }

%%%
%%% Stuff for bold maths typesetting  ----------------------------------------
%%%
%%%   e.g. use "\bfmu" for boldface mu symbol
%%%
\def\bfmu{\mbox{\boldmath$\mu$}}
\def\bftau{\mbox{\boldmath$\tau$}}
\def\bftheta{\mbox{\boldmath$\theta$}}
\def\bfdelta{\mbox{\boldmath$\delta$}}
\def\bfphi{\mbox{\boldmath$\phi$}}
\def\bfpsi{\mbox{\boldmath$\psi$}}
\def\bfeta{\mbox{\boldmath$\eta$}}
\def\bfnabla{\mbox{\boldmath$\nabla$}}
\def\bfGamma{\mbox{\boldmath$\Gamma$}}

%%% Make figure placement a little more predictable.
% We trust the user to move figures if this results
% in ugliness.
% Minimize bad page breaks at figures
%\renewcommand{\textfraction}{0.01}
%\renewcommand{\floatpagefraction}{0.99}
%\renewcommand{\topfraction}{0.99}
%\renewcommand{\bottomfraction}{0.99}
%\renewcommand{\dblfloatpagefraction}{0.99}
%\renewcommand{\dbltopfraction}{0.99}
%\setcounter{totalnumber}{99}
%\setcounter{topnumber}{99}
%\setcounter{bottomnumber}{99}
%
%% Add a period to the end of an abbreviation unless there's one
%% already, then \xspace.
%\makeatletter
%\DeclareRobustCommand\onedot{\futurelet\@let@token\@onedot}
%\def\@onedot{\ifx\@let@token.\else.\null\fi\xspace}
%
%\def\eg{\emph{e.g}\onedot} \def\Eg{\emph{E.g}\onedot}
%\def\ie{\emph{i.e}\onedot} \def\Ie{\emph{I.e}\onedot}
%\def\cf{\emph{c.f}\onedot} \def\Cf{\emph{C.f}\onedot}
%\def\etc{\emph{etc}\onedot} \def\vs{\emph{vs}\onedot}
%\def\wrt{w.r.t\onedot} \def\dof{d.o.f\onedot}
%\def\etal{\emph{et al}\onedot}
%\makeatother

% ---------------------------------------------------------------

\newcommand{\R}{\mathbb{R}}

\title{AdaptiveFusion: Adaptive Multi-Modal Multi-View Fusion for 3D Human Body Reconstruction}

\author{Anjun Chen, Xiangyu Wang, Zhi Xu, Kun Shi, Yan Qin, Yuchi Huo, Jiming Chen, \IEEEmembership{Fellow, IEEE,} and Qi Ye% <-this % stops a space
\IEEEcompsocitemizethanks{\IEEEcompsocthanksitem Anjun Chen, Xiangyu Wang, Zhi Xu, Kun Shi, Jiming Chen, and Qi Ye are with the State Key Laboratory of Industrial Control Technology, Zhejiang University. Email: \{anjunchen, xy\_wong, xuzhi, kuns, cjm, qi.ye\}@zju.edu.cn

Yan Qin is with the School of Automation, Chongqing University. Email: yan.qin@cqu.edu.cn

Yuchi Huo is with the State Key Lab of CAD\&CG, Zhejiang University and Zhejiang Lab. Email: eehyc0@zju.edu.cn

This work was supported in part by NSFC under Grants 62088101, 62233013, 61790571, 62103372, and the Fundamental Research Funds for the Central Universities. 
}}

% The paper headers
% \markboth{IEEE TRANSACTIONS ON PATTERN ANALYSIS AND MACHINE INTELLIGENCE,~2024}%
% The paper headers
\markboth{IEEE TRANSACTIONS ON MULTIMEDIA,~2025}%
{Shell \MakeLowercase{\textit{et al.}}: AdaptiveFusion: Adaptive Multi-Modal Multi-View Fusion for 3D Human Body Reconstruction}

% \IEEEpubid{0000--0000/00\$00.00~\copyright~2021 IEEE}
% Remember, if you use this you must call \IEEEpubidadjcol in the second
% column for its text to clear the IEEEpubid mark.

\maketitle

\begin{abstract}
Recent advancements in sensor technology and deep learning have led to significant progress in 3D human body reconstruction. However, most existing approaches rely on data from a specific sensor, which can be unreliable due to the inherent limitations of individual sensing modalities. \add{Additionally}, existing multi-modal fusion methods generally require customized designs based on the specific sensor combinations or setups, which limits the flexibility and generality of these methods. Furthermore, conventional point-image projection-based and Transformer-based fusion networks are susceptible to the influence of noisy modalities and sensor poses. To address these limitations and achieve robust 3D human body reconstruction in various conditions, we propose AdaptiveFusion, a generic adaptive multi-modal multi-view fusion framework that can effectively incorporate arbitrary combinations of uncalibrated sensor inputs. By treating different modalities from various viewpoints as equal tokens, and our handcrafted modality sampling module by leveraging the inherent flexibility of Transformer models, AdaptiveFusion is able to cope with arbitrary numbers of inputs and accommodate noisy modalities with only a single training network. Extensive experiments on large-scale human datasets demonstrate the effectiveness of AdaptiveFusion in achieving high-quality 3D human body reconstruction in various environments. In addition, our method achieves superior accuracy compared to state-of-the-art fusion methods.
\end{abstract}

% Note that keywords are not normally used for peerreview papers.
\begin{IEEEkeywords}
3D human body reconstruction, adaptive multi-modal multi-view fusion.
\end{IEEEkeywords}

\section{Introduction}
\label{sec:intro}
3D human body reconstruction is widely studied in many practical vision applications, including human-robot interaction, XR technologies, and motion capture. Most of the prevailing reconstruction approaches leverage single-modal data, such as RGB image \cite{kanazawa2018end,lin2021end,feng2022fof}, depth image \cite{hu20213dbodynet,liu2021votehmr}, and radar point cloud \cite{xue2021mmmesh,chen2022mmbody}.
Nonetheless, all kinds of sensing modalities have their own defects, and therefore single-modal information sources inevitably suffer from unreliability. For example, the perception capability of the RGB camera deteriorates rapidly in poor illumination and inclement weather \cite{bijelic2020seeing}. Radar signals are restricted by sparsity and multi-path effect \cite{shi2021road}, while depth camera performs poorly in smoke and occlusion conditions \cite{zhang2021sequential}. Therefore, fusing sensory data from disparate modalities to combine their strengths is essential to realize robust 3D human body reconstruction in both normal and adverse conditions.

\begin{table}[t!]
  \centering
  \caption{\add{Results (cm)} of different fusion methods for RGB images and depth (and noisy) point clouds on the mmBody dataset \cite{chen2022mmbody}. Noisy Kinect depth is downsampled Kinect depth point clouds with missing parts.}
  \resizebox{0.48\textwidth}{!}{
    \begin{tabular}{c|cc|cc}
    \toprule
    \multirow{2}[4]{*}{Methods} & \multicolumn{2}{c|}{Image w/ Kinect Depth} & \multicolumn{2}{c}{Image w/ Noisy Kinect Depth} \\
\cmidrule{2-5}          & MPJPE ↓ &  MPVE ↓ & MPJPE ↓ &  MPVE ↓ \\
    \midrule
    Point w/ Image Feature \cite{wang2021pointaugmenting} & 4.4   & 5.8   & \underline{5.0} (\underline{13\%} ↑)  & \underline{6.5} (\underline{12\%} ↑)\\
    DeepFusion \cite{li2022deepfusion} & \underline{4.3}   & \underline{5.5}   & 5.3 (23\% ↑) & 6.6 (20\% ↑)\\
    TokenFusion \cite{wang2022multimodal} & 4.7   & 6.2   & 7.8 (65\% ↑)  & 10.4 (67\% ↑)\\
    AdaptiveFusion (Ours) & \textbf{3.5} & \textbf{4.7} & \textbf{3.7} (\textbf{5\%} ↑)& \textbf{4.9} (\textbf{4\%} ↑)\\
    \bottomrule
    \end{tabular}%
    }
  \label{tab:sparsity}%
\end{table}%

However, combining multi-modal information is not trivial. Early LiDAR-camera fusion approaches \cite{vora2020pointpainting,wang2021pointaugmenting} adopt point-to-image projection to combine point clouds and image pixel values/features through element-wise addition or channel-wise concatenation. These approaches rely on the local projection relationship between the point clouds and images, which can break down if one of the modalities is compromised or fails. Undesirable issues like random incompleteness and temporal fluctuations of point clouds can result in the retrieval of inadequate or incorrect features from corresponding images \cite{chen2023immfusion}. Furthermore, the degradation of image features in challenging environments, such as low lighting conditions, can extremely impair the performance \cite{shi2024radar}. More recently, several customized Transformer-based structures~\cite{li2022deepfusion,wang2022multimodal,wu2023transformer} have been proposed for multi-modal fusion. These networks still suffer from the issues of noisy point clouds. As demonstrated in \cref{tab:sparsity}, both point-image projection-based and transformer-based fusion networks deteriorate rapidly with noisy depth\footnote{\chen{We process the original human depth point clouds to generate noisy depth point clouds with missing parts and sparsity. Specifically, we utilize ground-truth joint locations to remove most of the depth points in the lower body region of the human body, and then randomly remove points near 1-5 limb joints to simulate the random missing characteristics of the radar point cloud. Subsequently, the remaining depth point cloud is downsampled to 256 points and fed into the fusion models with RGB images.}}.

Besides, most of these fusion frameworks work with the fusion of two modalities from the same viewpoint. Only several recent works \cite{chen2023futr3d,wang2023unitr} explore flexible sensor fusion for object detection. These works are constrained by the requirements of different training weights for various modal combinations or accurate poses for different sensors and views. Despite various fusion methods, to the best of our knowledge, how to integrate information from diverse uncalibrated sensor setups and combinations in an adaptive way remains unexplored. With the development of mobile robots and the Internet of Things, information fusion from dynamically combined multi-view sensors with unknown/noisy poses is an important problem to be solved to enable collaboration between agents.

Therefore, in this paper, we present AdaptiveFusion, the first generic multi-view and multi-modal fusion framework for 3D human body reconstruction that can adaptively fuse arbitrary numbers of uncalibrated sensor inputs from different viewpoints to achieve flexible and robust reconstruction. 
\modi{Most existing fusion methods primarily focus on point cloud as the main modality to be merged into, which 1) cannot handle arbitrary numbers and orders of modalities, 2) will deteriorate rapidly if the point clouds become sparse and noisy, and 3) cannot deal with data without sensor calibration.}
In our framework, different from fusion via projection, we resort to a Fusion Transformer Module to adaptively select local features from different inputs based on their feature strengths instead of the spatial affinity of features and fuse the more informative features with quantity-irrelevant operators. Additionally, in contrast to previous fusion methods that regard point clouds as the main modality, our framework does not assume a main modality and treats features from different modalities as equal tokens. The corrupted tokens from one modality could possibly be remedied by others or disregarded to accommodate the sparsity and missing parts. 
Further, we propose an innovative modality sampling module, ensuring the model encounters all modal combinations during training. Consequently, we only need to train a single network to handle arbitrary input combinations during inference.
% Further, AdaptiveFusion utilizes general backbones to extract global and local features from various modalities to address the issue of feature misalignment and to reinforce the interactions between global and local contexts. 
With all designs, \ie the notion of equal tokens for different modalities rather than a predefined main modality and the ingenious training strategy, our framework not only enables to accommodate arbitrary numbers of modalities but also enhances fusion capability to handle the fusion with noisy point clouds much more effectively as verified in \cref{tab:sparsity}. 

We conduct extensive experiments on the mmBody \cite{chen2022mmbody} dataset, including extreme weather conditions like smoke, rain, and night. We evaluate the performance of AdaptiveFusion under different scenes, and it outperforms traditional fusion methods in all weather environments. Additionally, we evaluate AdaptiveFusion on the other large-scale human datasets Human3.6M \cite{ionescu2013human3}, HuMMan \cite{cai2022humman}, BEHAVE \cite{bhatnagar22behave}, and 3DPW \cite{von2018recovering}, and it outperforms existing works by a large margin.
The contributions of our work can be summarized as follows: 
\begin{itemize}
\item We propose AdaptiveFusion, the first generic adaptive fusion framework that can adapt to arbitrary combinations of multi-modal multi-view inputs without calibration for 3D human body reconstruction. 
\item Our novel notion of treating different modalities as equal tokens and the incorporation of coupled attention modules effectively handles fusion with noisy point clouds while also enabling a generic framework. 
\item AdaptiveFusion realizes excellent performance across different sensor configurations, is insusceptible to sensor defects in various weather conditions, and outperforms stat-of-the-art fusion approaches significantly. 
\end{itemize}

This paper extends our previous conference version \cite{chen2023immfusion} of fusing two fixed modalities from the same viewpoint to fusing arbitrary combinations of multi-modal multi-view inputs. As the current submission addresses a different problem, it only keeps the idea of fusion by attention operations on tokens from different modalities while the introduction, the related work, the detailed methodology, and the experiments all differ from the conference version. 
% \chen{This paper is an extension of our conference version \cite{chen2023immfusion}. While sharing the similarity in methodology, this paper extends the last version in the following aspects: First, we generalize our method from dual-modal input to adaptively accommodate arbitrary modalities and viewpoints without calibration. Second, we conduct comprehensive experiments across single-view, multi-view, and multi-modal datasets, validating our method can adapt to dynamical multi-modal multi-view inputs effectively. Third, we benchmark our method against the previous version and demonstrate its superior performance across all scenarios.}
The rest of this paper is organized as follows: \cref{sec:rw} gives a brief overview of related works on 3D human reconstruction and multi-sensor fusion. \cref{sec:method} presents our proposed adaptive multi-modal multi-view fusion method, AdaptiveFusion. \cref{sec:exp} elaborates experimental results. \cref{sec:con} finally concludes the paper.

\section{Related Works}
\label{sec:rw}
\subsection{Human Body Reconstruction}

3D human body reconstruction can be broadly categorized into parametric \cite{kanazawa2018end,kolotouros2019learning,li2022cliff} and non-parametric approaches \cite{kolotouros2019convolutional,lin2021end,kim2023sampling}. In the former, a body model, such as skeleton \cite{huang2020generating,liu2023posynda}, SMPL \cite{loper2015smpl}, and SMPL-X \cite{pavlakos2019expressive}, is used to represent the human body. Despite greatly reducing regressing parameters, it is still challenging to estimate precise coefficients from a single image\cite{pavlakos2022human}. To improve the reconstruction, researchers make efforts by utilizing multi-view information \cite{zhao2018multi,liang2019shape,kolotouros2021probabilistic,jia2023paff}, temporal sequences\cite{kocabas2020vibe,tu2024motionfollower,tu2024motioneditor}, or dense depth maps \cite{zhao20183,xu2020building,liu2021votehmr,hu20213dbodynet}. On the other hand, non-parametric approaches directly regress the vertices of the 3D mesh from the input image. Most pioneers choose Graph Convolutional Neural Network \cite{kolotouros2019convolutional} to model the local interactions between neighboring vertices with an adjacency matrix. More recent approaches, such as METRO \cite{lin2021end} and Graphormer \cite{lin2021mesh}, utilize transformer encoders to jointly model the relationships between vertices and joints. 
Recently, millimeter wave (mmWave) sensors have gained popularity for their ability to work in challenging conditions such as rain, smoke, and occlusion. Several wireless systems \cite{zhao2018through,wang2019person,yu2023mobirfpose} have been developed to reconstruct the human body and the mmWave-based system is one of them. Xue et al. \cite{xue2021mmmesh} propose an accessible real-time human mesh reconstruction solution utilizing commercial portable mmWave devices. 
Chen et al. \cite{chen2022mmbody} present a large-scale mmWave human body dataset with paired RGBD images in various environments. With this dataset, Chen et al. \cite{chen2023immfusion} introduce an mmWave-RGB fusion method for 3D human body reconstruction. These methods have shown decent results in 3D human body reconstruction from various modalities. However, the capability of the reconstruction from the uncalibrated multi-sensor combinations is not studied. In this work, we adopt the non-parametric approach for body reconstruction from unfixed sensor inputs. 

% Among the works mentioned above, Immfusion \cite{chen2023immfusion} is the most relevant work to our proposed method. The main difference is that we are designing a generic adaptive fusion method with arbitrary modality combinations for human body reconstruction. 

% Most prevalent work is dedicated to modalities associated with images (RGB, infrared) and point clouds (radar, LiDAR). 
\subsection{Multi-Sensor Fusion}
% AI-driven multi-sensor fusion roughly falls into three categories, namely data, feature, and decision levels. 
% Generally, how to overcome disparateness and exploit the synergy of heterogeneous modalities are the foremost considerations. To this end, 
% resort to investigating elaborately designed modal alignment schemes. 
% Conventional fusion methods can be broadly classified into three categories: data-level, feature-level, and decision-level fusion. Data-level fusion  \cite{cheng2021robust,long2021radar} commonly entails coordinate projection technique. Decision-level fusion \cite{nabati2021centerfusion} usually utilizes information from one modality to generate regions of interest containing valid objects. Feature-level fusion \cite{kim2020grif} typically involves the fusion of proposal-wise features in multi-view maps. All these conventional approaches make efforts in CNN-based modal alignment schemes, while most of them are short of efficiency, adaptability, and compatibility.
Conventional fusion methods can be broadly classified into three categories: decision-level, data-level, and feature-level fusion. 
Decision-level fusion \cite{chen2017multi,qi2018frustum,nabati2021centerfusion} usually utilizes information from one modality to generate regions of interest containing valid objects. 
However, such coarse-grained fusion strategies may not fully release the potential of multiple modalities. 
Data-level fusion  \cite{vora2020pointpainting,cheng2021robust,long2021radar,wang2021pointaugmenting} commonly entails the coordinate projection technique, which is easily affected by the sensor misalignment and defective image features. 
Feature-level fusion \cite{huang2020epnet,zhou2020novel,piergiovanni20214d,wan2023mffnet} typically involves the fusion of proposal-wise features in multi-modal feature maps, while determining the optimal weighting for features of each modality is challenging. 
% All these conventional approaches make efforts in modal alignment schemes, while most of them are short of efficiency, adaptability, and compatibility.
Recently, promising performance has been achieved by Transformer-based fusion, which sheds light on the possibility of leveraging the Transformer structure as a substitute for manually designed alignment operations. 
Specifically, DeepFusion \cite{li2022deepfusion} uses a learnable alignment mechanism to dynamically correlate LiDAR information with the most relevant camera features.
TokenFusion \cite{wang2022multimodal} prunes feature tokens among single-modal Transformer layers to preserve better information and then re-utilizes the pruned tokens for multi-modal fusion. 
TransFusion \cite{bai2022transfusion} employs a soft-association approach to process inferior image situations. 
Chen \etal ~\cite{chen2023immfusion} propose a radar-camera fusion method ImmFusion to combine mmWave point clouds and RGB images to reconstruct 3D human body. 
However, these methods have not achieved the capability of handling unfixed numbers of input modalities.

Currently, only \yq{limited} literature reveals insights on fusion methods with more than three sensors. SeeingThroughFog \cite{bijelic2020seeing} advances an entropy-driven fusion architecture utilizing RGB cameras, LiDARs, radars, and infrared cameras for object detection. Some recent works \cite{chen2023futr3d,yan2023cross,wang2023unitr} propose to formulate unified end-to-end multi-sensor fusion frameworks for 3D detection. 
These multi-modal fusion schemes, however, can not handle arbitrary modalities without calibration and are tailored for the object detection task, which is inapplicable for the fusion-based human body reconstruction task. To this end, we make efforts in overcoming potential sensor defects to robustly reconstruct 3D human body from multi-view multi-modal inputs in various weather conditions.

\section{Method}
\label{sec:method}
\begin{figure*}
\centering
    \includegraphics[width=\linewidth, trim=0 0 0 0, clip]{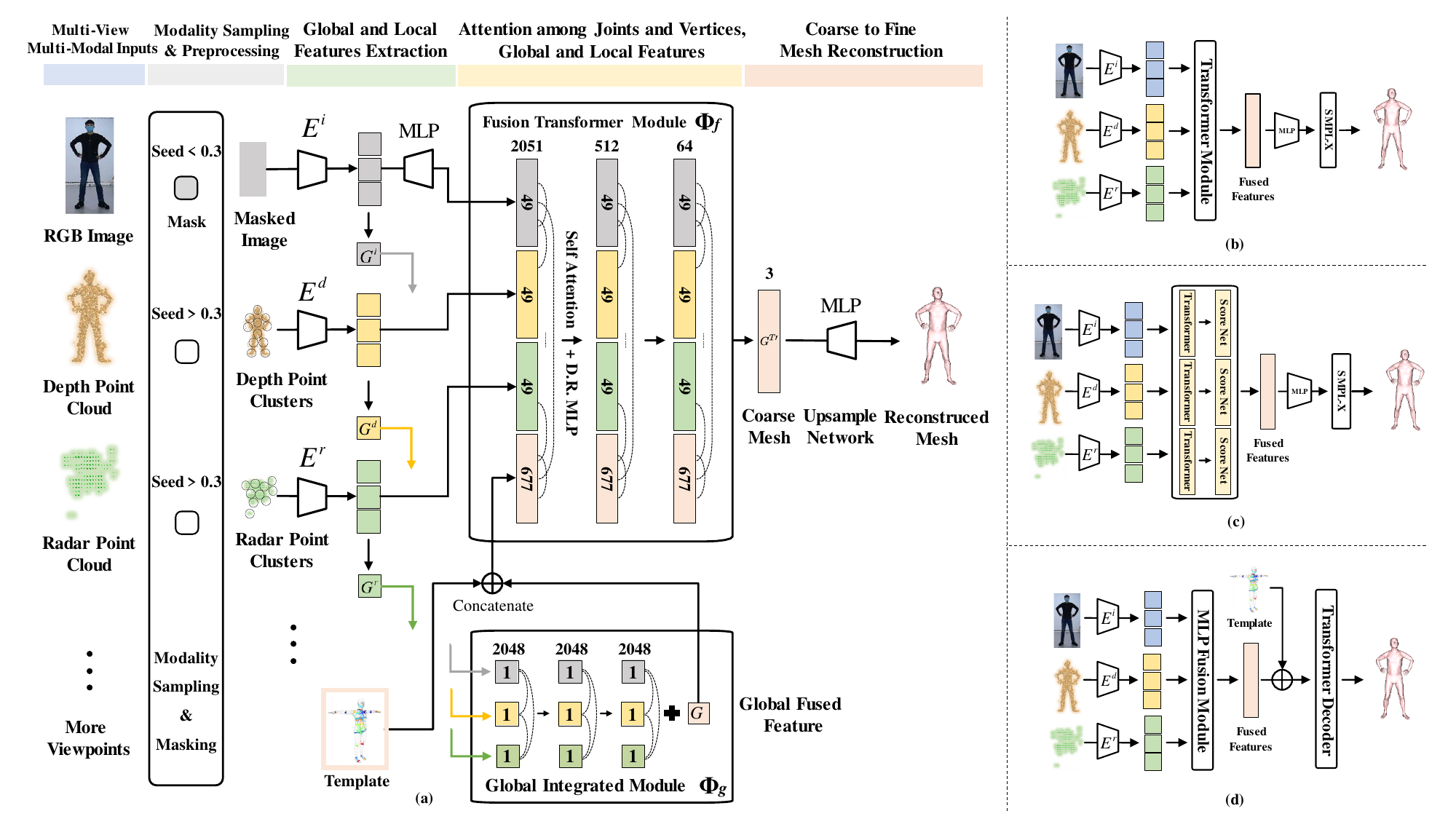}
    \caption{Comparison of different fusion methods. (a) Framework of our proposed AdaptiveFusion. 
    We first extract global and local features from each of the sampled modalities using corresponding backbones. Next, we utilize Global Integrated Module to incorporate global features. Then, we employ Fusion Transformer Module to fuse global and local features and to regress locations of joints and vertices. 
    D.R. MLP stands for a dimension reduction MLP. 
    (b) DeepFusion \cite{li2022deepfusion}. (c) TokenFusion \cite{wang2022multimodal}. (d) FUTR3D \cite{chen2023futr3d}
    }
    \label{fig:framework}
\end{figure*}

In this section, we present our proposed AdaptiveFusion, which takes arbitrary numbers of uncalibrated sensor inputs for 3D human body reconstruction.
\cref{fig:framework} illustrates the framework of AdaptiveFusion. The structure aims to effectively utilize the sensor information at global and local levels to predict the human body mesh. Given any sensor inputs, the global/local features for each modality are firstly extracted by the corresponding backbone, respectively. Next, global features of different modalities are incorporated as one global feature vector by the Global Integrated Module (GIM) and embedded with SMPL-X template positions. Then, all global/local features are tokenized as input of a multi-layer Fusion Transformer Module (FTM) to dynamically fuse the information of all modalities and directly regress the coordinates of 3D human joints and coarse mesh vertices. \yq{With} the adaptability to the arbitrary number of token inputs of Transformer, GIM and FTM can adaptively handle any input feature combinations, including single modality input. At last, we employ Multi-Layer Perceptrons (MLPs) to upsample the coarse mesh vertices to the full mesh vertices. 

\subsection{Problem Formulation}
\label{Preliminary}

AdaptiveFusion aims to predict the 3D positions of the joints and vertices of human meshes from any input modality combinations:
\begin{equation}
f:s\rightarrow Y \quad s \in S(X),
\end{equation}
where $X$ is a set of different input modalities, $S(X)$ contains all permutations of the nonempty power set of $X$, $s$ denotes one combination of input sensors, and $Y$ is the output joints and vertices.

Specifically, the available inputs of the mmBody dataset \cite{chen2022mmbody} are RGB images and depth point clouds from 2 viewpoints, and radar point clouds from one viewpoint. Hence, $X=\{I_{m}, D_{m}, R\}, m=1,2$. $I_{m}\in \mathbb{R}^{224 \times 224 \times 3}$ is the cropped body region of RGB images with a size of $224\times224$,  $D_{m}\in \mathbb{R}^{4096 \times 3}$ depth point clouds with 4096 points, and $R\in \mathbb{R}^{1024 \times 3}$ radar point clouds with 1024 points. $m$ indicates the viewpoints. With these inputs, a total of 31 input combinations are supported. The output $Y=[J, V]$, where $J \in \mathbb{R}^{22 \times 3}, V\in \mathbb{R}^{10475 \times 3}$ are the XYZ coordinates of 22 joints and 10475 vertices. The number of joints and vertices varies depending on the body template used, and the template in the mmBody dataset is SMPL-X \cite{pavlakos2019expressive} template. 
As we focus on reconstruction, we utilize the bounding boxes automatically annotated from the ground-truth mesh joints to crop the region of interest containing only the body part. 

% For brevity, we only present full modalities as input here, and the other modality combinations will be covered in the experimental section. Therefore, we leave out the superscript $t$ in the following parts.

% where $I_{m}^{t}\in \mathbb{R}^{224 \times 224 \times 3}, D_{m}^{t}\in \mathbb{R}^{4096 \times 3}, R^{t}\in \mathbb{R}^{1024 \times 3}$ are the cropped body region of RGB images with a size of $224\times224$, depth point clouds with 4096 points, radar point clouds with 1024 points, and $J^{t} \in \mathbb{R}^{22 \times 3}, V^{t}\in \mathbb{R}^{10475 \times 3}$ are the predicted locations of 22 joints and 10475 vertices at time $t$. As our task is reconstruction, we utilize the bounding boxes automatically annotated from the ground-truth mesh joints to crop the region of interest containing only the body part. Each frame includes a random sample of different input modalities, which are then combined and fed into the network. In this work, $M = 2$, and the training set contains a total of 32 modality combinations as input. For brevity, we only present full modalities as input here, and the other modality combinations will be covered in the experimental section. Therefore, we leave out the superscript $t$ in the following parts.

\subsection{Muilti-Modal Multi-View Feature Extraction}

In order to address the issue of feature alignment, we independently extract global and local features for each modality following \cite{chen2023immfusion}. This strategy allows for the better extraction of global contextual dependencies and modeling of local interactions. Specifically, we directly feed the human body region of multi-modal inputs to the commonly used backbones to extract features. Backbones can be replaced with alternative options for new modalities input.

We use a shared HRNet \cite{wang2020deep} $E^{i}$ to acquire \yq{local} image grid features $L_{m}^{i}\in \mathbb{R}^{49\times 2051}$ and the global image feature $G_{m}^{i}\in \mathbb{R}^{2048}$ from multi-view images $I_{m}$. Then we use an MLP to reduce the dimension of grid features
to the same size as the cluster features of point clouds.
For multi-view depth point clouds $D_{m}$, we obtain \yq{local} depth cluster features $L_{m}^{d}\in \mathbb{R}^{49\times(3+2048)}$ using shared PointNet++ \cite{qi2017pointnet++} $E^{d}$, where 49 denotes the number of seed points sampled by Farthest Point Sample (FPS), 3 denotes the spatial coordinate, and 2048 denotes the dimension of features extracted from the grouping local points. A global feature vector $G_{m}^{d}\in \mathbb{R}^{2048}$ is further extracted from cluster features $L_{m}^{d}$ using an MLP. 
We employ another PointNet++ $E^{r}$ to extract local radar cluster features $L^{r}$ and global feature $G^{r}$ for the radar point cloud $R$ due to its sparse and noisy nature.

\subsection{Modality Fusion with Transformers}

Multi-head attention module \cite{vaswani2017attention} has been effective for modeling the relationship between information tokens and processing unordered, heterogeneous, and length-undefined data structures such as words and sentences. To allow our fusion framework to effectively select informative token features from arbitrary input modalities, and to dynamically fuse these features from different viewpoints and different modalities, we formulate our fusion problem into the attention framework by exacting \textit{words} (local features) and \textit{sentences} (global features) from different inputs and designing the interaction modules between these \textit{words} and \textit{sentences}.
\modi{Given the input modality tokens $\mathbf{X} = { \mathbf{x}_1, \mathbf{x}_2, \dots, \mathbf{x}_n }$, the tokens are first projected into queries $\mathbf{Q}$, keys $\mathbf{K}$, and values $\mathbf{V}$ using trainable parameters ${ \mathbf{W}_Q, \mathbf{W}_K, \mathbf{W}_V }$:
\begin{equation}
    \mathbf{Q}, \mathbf{K}, \mathbf{V} = \mathbf{X} \mathbf{W}_Q, \mathbf{X} \mathbf{W}_K, \mathbf{X} \mathbf{W}_V.
\end{equation}
The output $\mathbf{Y} = { \mathbf{y}_1, \mathbf{y}_2, \dots, \mathbf{y}_n }$ is then computed as:
\begin{equation}
    \mathbf{Y} = \text{Att}(\mathbf{Q}, \mathbf{K}, \mathbf{V}),
\end{equation}
where $\text{Att}(\cdot)$ represents the attention function, which determines the semantic relevance of a query $\mathbf{Q}$ to the keys $\mathbf{K}$ using scaled dot-product and softmax operations.
When a modality, such as depth, encounters occlusion or noise, it will generate a higher loss, which is back propagated to $\{ \mathbf{W}_Q, \mathbf{W}_K, \mathbf{W}_V \}$ through the gradient. This process allows the attention module to dynamically adjust the weight of each modality, learning to prioritize more reliable information from image or radar modalities. The attention function $\text{Att}(\cdot)$ computes weights \(w_1, w_2, w_3, \dots\) for each modality, and adapts these weights based on the relative reliability of the features.}
% Given $n$ modality tokens $m_{1}, m_{2}, ..., m_{n}$, our fusion module updates vertices $V$ (also tokens) by aggregating these features as $V=w_{1}m_{1}+w_{2}m_{2}+...+w_{n}m_{n}$, where the weight is determined by the similarity of tokens. With the invariance of $w_{1}, w_{2}, ..., w_{n}$ to the order and quantity, our framework can incorporate arbitrary combinations of sensor inputs. For the noisy modalities $m_{k}$, the model can learn to decrease their weights $w_{k}$ because noisy tokens are far from normal distribution and their similarities with other tokens are lower.

% The self-attention mechanism allows the model to effectively select informative token features from arbitrary input modalities, and to dynamically fuse these features. 
% We adopt this structure to mitigate the feature degradation caused by the sparsity and noise of mmWave signals and the deficiency of RGB or depth information in extreme conditions for full-modalities input. Simultaneously, it is adaptable to any modality changes. 

Specifically, all global features are firstly fused into a global feature $G \in \mathbb{R}^{2048}$ by Global Integrated Module (GIM) $\Phi_g$ implemented using a tiny Transformer module similar to \cite{chen2023immfusion},
\begin{equation}
 G =\Phi_g (G_{m}^{i}, G_{m}^{d}, G^{r}),
\end{equation}
where $\Phi_g$ is a three-layer attention module ending with a sum operation to integrate the global features. The fusion of global features provides overall context information to mitigate the feature degradation caused by sparsity,  missing parts, and adverse weather conditions.

After $\Phi_g$, we perform positional embedding by attaching the 3D coordinates of 22 joints and 655 vertices in a coarse mesh downsampled from a SMPL-X template mesh to the global vector, $G^{T} = cat(J^{template}, V^{template}, G)$, where $G^{T} \in \mathbb{R}^{677 \times 2051}$. These template features serve like initial object queries in AnchorDETR \cite{wang2022anchor} for 3D object detection.
Simultaneously, the ordinal numbers of each modality are embedded in local features, \ie cluster and grid features. The incorporation of these local features enables remedy or discard for a local corrupted area only, rather than the entire modality in order to retain as much effective information as possible.

% Simultaneously, the ordinal numbers of each modality are embedded in local features, \ie cluster and grid features, to simplify the training. All local features serve the purpose of providing fine-grained local details for body reconstruction and 

Second, we utilize the Fusion Transformer Module (FTM) $\Phi_f$ to adaptively fuse the information from arbitrary inputs. FTM transforms all modality features into Queries and conducts self-attention between them:
\begin{equation}
\begin{split}
G^{T\prime}, L_{m}^{i\prime}, L_{m}^{d\prime}, L^{r\prime}=\Phi_{f}\left(G^{T}, L_{m}^{i}, L_{m}^{d}, L^{r}\right),
\end{split}
\end{equation}
where coarse mesh $G^{T\prime}\in \mathbb{R}^{677 \times 3}$, $L_{m}^{i\prime}$, $L_{m}^{d\prime}$, and $L^{r\prime}\in \mathbb{R}^{49 \times 3}$. $\Phi_{f}$ is implemented with a three-layer Transformer module that uses several attention heads in parallel to fuse global and local features. 
While attending to valid features and restricting undesirable features, FTM $\Phi_f$ adopts attention between joint/vertex queries $G^{T}$ generated from global features $G$ and modality tokens from local features $L_{m}^{i}, L_{m}^{d}, L^{r}$ to aggregate relevant contextual information for multi-modal input. Additionally, the self-attention mechanism helps reason interrelations between each pair of candidate queries for single-modal input. Then, we adopt a dimension-reduction architecture, Graph Convolution \cite{kolotouros2019convolutional}, to decode the queries $G^{T\prime}$ containing rich cross-modalities information into 3D coordinates of joints and vertices following \cite{lin2021mesh}. Last, a linear projection network implemented using MLPs upsamples the coarse output mesh to the original 10475 vertices.

\subsection{Modality Sampling and Masking}
To make our model more versatile, we design a modality sampling module to randomly sample one combination of input modalities during a forward training process. This simple yet effective strategy ensures the model encounters all possible combinations of modalities within a single training session, enabling the model to adapt to different modality combinations without retraining. Besides, we empirically observe that our strategy surprisingly enables one combination to benefit from other combinations: even with a small portion of training data for each modality combination, our model achieves better results than the performance of models training with fixed modality as demonstrated in \cref{fig:mpjpe}. For the current era of large models where training often involves datasets on the scale of millions, this discovery is highly advantageous.

Despite the superiority of the multi-head attention mechanism, the model is prone to struggle with data imbalance of training data (without data under adverse conditions) for multi-modal input according to \cite{bijelic2020seeing}, which makes the Transformer focus all attention on the single modality that performs better under normal circumstances like image or depth data. 
Similar to \cite{chen2023immfusion}, to effectively activate the potential of the model across all scenarios,
we utilize a modality masking module to randomly mask some of the modalities for multi-modal input and thus enforce the model to learn from all modalities in various situations. 
As a result, this module enables our model to overcome the training data bias problem and consider all modalities, which further facilitates the model to perform better across all scenarios in our experiments. 
For the mask proportion, we set it to 30\% in our experiments as it achieves the best accuracy.

\begin{table*}[ht!]
  \centering
  \caption{Mean errors (cm) of different methods for 3D body reconstruction in different scenes of mmBody dataset \cite{chen2022mmbody}. For the two columns of each scene, the first column is for MPJPE and the second is for MPVE.}
  \resizebox{\textwidth}{!}{
    \begin{tabular}{c|cc|cc|cc|cc|cc|cc|cc|cc}
    \toprule
    \multicolumn{1}{c|}{\multirow{2}[2]{*}{Methods}} & \multicolumn{6}{c|}{Basic Scenes}             & \multicolumn{8}{c|}{Adverse Environments}                     & \multicolumn{2}{c}{\multirow{2}[2]{*}{Average}} \\
          & \multicolumn{2}{c}{\textbf{Lab1}} & \multicolumn{2}{c}{\textbf{Lab2}} & \multicolumn{2}{c|}{\textbf{Furnished}} & \multicolumn{2}{c}{\textbf{Rain}} & \multicolumn{2}{c}{\textbf{Smoke}} & \multicolumn{2}{c}{\textbf{Poor Lighting}} & \multicolumn{2}{c|}{\textbf{Occlusion}} & \multicolumn{2}{c}{} \\
    \midrule
    % Image-Only & 4.1   & 5.5   & 4.0   & 5.3   & 5.4   & 6.8   & 5.9   & 7.4   & 8.5   & 11.2  & 9.9   & 14.1  & 11.3  & 16.6  & 7.0   & 9.6 \\
    % Depth-Only & 3.5   & 4.7   & 3.8   & 5.0   & 4.6   & 5.4   & 7.0   & 8.2   & 12.3  & 15.6  & 4.5   & 5.7   & 12.5  & 16.3  & 6.9   & 8.7 \\
    % Radar-Only & 6.3   & 8.8   & 6.7   & 8.9   & 6.4   & 8.8   & 7.8   & 10.2  & 8.0   & 10.6  & 6.2   & 8.4   & 8.8   & 12.7  & 7.2   & 9.8 \\
    DeepFusion \cite{li2022deepfusion} & 4.3   & 5.4   & 4.1   & 5.3   & 5.6   & 7.2   & 5.0   & 6.3   & 7.6   & 9.3   & 5.2   & 6.2   & 7.8   & 10.5  & 5.7   & 7.2 \\
    % DeepFusion (non-parametric) & 4.1   & 5.2   & 4.0   & 5.2   & 4.8   & 6.0   & 4.5   & 5.6   & 7.7   & 10.1   & 4.5   & 5.8   & 7.2   & 9.5  & 5.3   & 6.8 \\
    TokenFusion \cite{wang2022multimodal} & 4.4   & 5.9   & 4.5   & 6.0   & 5.3   & 7.0   & 5.1   & 6.6   & 8.2   & 10.6  & 9.4   & 13.6  & 9.2   & 12.9  & 6.6   & 8.9 \\
    % TokenFusion (non-parametric) & 4.5   & 5.9   & 4.4   & 5.7   & 6.5   & 8.6   & 5.0   & 6.2   & 6.9   & 8.8   & 9.3   & 13.4  & 9.7   & 13.6   & 6.6   & 8.9  \\
    FUTR3D \cite{chen2023futr3d} & 4.0   & 5.3   & 4.1   & 5.3   & 4.7   & 6.0   & \underline{4.3}   & \underline{5.4}   & 7.8   & 10.2   & 4.4   & 5.7   & \underline{7.2}   & \underline{9.5}  & \underline{5.3}   & \underline{6.8} \\
    ImmFusion \cite{chen2023immfusion} & 4.1   & 5.4   & 3.7   & 4.7   & 5.2   & 6.4   & 5.6   & 6.8   & 7.6   & 9.8   & 6.8   & 9.0   & 7.8   & 11.0  & 5.9   & 7.4 \\
    \midrule
    AdaptiveFusion (img1) & 4.3   & 6.0   & 4.5   & 5.9   & 5.9   & 7.3   & 5.5   & 6.5   & 8.6   & 11.6  & 9.8   & 13.8  & 9.9   & 13.7  & 6.9   & 9.2 \\
    AdaptiveFusion (dep1) & \underline{3.3}   & \underline{4.3}   & \underline{3.4}   & \underline{4.3}   & \underline{4.0}   & \underline{4.9}   & 5.3   & 6.5   & 10.6  & 15.0  & \underline{3.5}   & \underline{4.4}   & 9.4   & 14.0  & 5.6   & 7.6 \\
    AdaptiveFusion (radar) & 6.1   & 8.1   & 5.5   & 7.4   & 5.9   & 8.1   & 7.0   & 8.8   & 8.1   & 10.1  & 5.8   & 7.9   & 8.0   & 10.7  & 6.7   & 8.7 \\
    AdaptiveFusion (img1-radar) & 4.1   & 5.4   & 4.2   & 5.4   & 5.2   & 6.3   & 5.2   & 6.0   & \underline{7.0}   & \underline{9.0}   & 6.4   & 8.6   & 7.7   & 10.1  & 5.8   & 7.3 \\
    % AdaptiveFusion-w/o-LF & 4.6   & 5.9   & 4.3   & 4.8   & 5.1   & 6.3   & 5.3   & 6.6   & 7.9   & 10.4  & 6.6   & 8.2   & 9.8   & 14.0  & 6.2   & 8.0 \\
    % AdaptiveFusion-w/o-GIM & 3.4   & 4.6   & 3.7   & 4.8   & 4.1   & 5.2   & 4.1   & 5.3   & 6.2   & 7.9   & 3.9   & 4.7   & 7.1   & 10.1  & 4.7   & 6.1 \\
    % AdaptiveFusion-w/o-MMM & 3.5   & 4.7   & 3.6   & 4.7   & 4.2   & 5.4   & 3.9   & 5.1   & 6.6   & 8.8   & 5.2   & 6.6   & 10.3  & 14.7  & 5.3   & 7.1 \\
    AdaptiveFusion (full-modalities) & \textbf{3.1} & \textbf{4.1} & \textbf{3.3} & \textbf{4.2} & \textbf{3.7} & \textbf{4.4} & \textbf{3.4} & \textbf{4.4} & \textbf{5.2} & \textbf{6.4} & \textbf{3.5} & \textbf{4.3} & \textbf{4.5} & \textbf{6.6} & \textbf{3.8} & \textbf{4.9} \\
    \bottomrule
    \end{tabular}%
    }
  \label{tab:errors}%
\end{table*}%

\subsection{Training Loss} 
% Our AdaptiveFusion applies $L_1$ loss to constrain the reconstructed 3D joints.
% \begin{equation}
% \mathcal{L}_J^{3D}=\sum_{i=1}^N\left|\left|J_{3D}-\bar{J}_{3D}\right|\right|_1.
% \end{equation}
% We also incorporate $L_1$ loss to the 2D joints obtained by projecting predicted 3D joints to the image space using estimated camera parameters.
% \begin{equation}
% \mathcal{L}_J^{2D}=\sum_{i=1}^N\left|\left|J_{2D}-\bar{J}_{2D}\right|\right|_1.
% \end{equation}
% In addition, the vertices of each layer of the Fusion Transformer Module are also supervised by ground truth meshes using $L_1$ loss.
% \begin{equation}
% \mathcal{L}_V=\sum_{i=1}^M\left|\left|V-\bar{V}\right|\right|_1.
% \end{equation}
Our AdaptiveFusion applies $L_1$ loss to the reconstructed mesh to constrain the 3D vertices $V$ and joints $J$. We also incorporate $L_1$ loss to the 2D joints $J_{2D}$ obtained by projecting predicted 3D joints to the image space using estimated camera parameters. In addition, the coarse meshes $V_{d1}, V_{d2}$ are also supervised by downsampled ground truth meshes using $L_1$ loss to accelerate convergence. 
The total loss of AdaptiveFusion is calculated by:
% \begin{equation}
% \mathcal{L}=\alpha\mathcal{L}_J^{3D} + \beta\mathcal{L}_J^{2D} + \gamma\mathcal{L}_V,
% \end{equation}
\begin{equation}
\begin{aligned}
\mathcal{L} = \alpha\|J-\bar{J}\|_1 + \beta\|J_{2D}&-\bar{J}_{2D}\|_1 + \gamma(\|V-\bar{V}\|_1 \\+ \|V_{d1}-\bar{V}_{d1}\|_1 &+ \|V_{d2}-\bar{V}_{d2}\|_1),
\end{aligned}
\end{equation}
where $\alpha$, $\beta$, and $\gamma$ denote the weight of each component, and variables with overline represent the ground truth.

\subsection{Comparison with Relevant Methods}
\label{sec:comparison}
We compare our proposed AdaptiveFusion with other relevant fusion methods, \ie DeepFusion \cite{li2022deepfusion}, TokenFusion \cite{wang2022multimodal}, and FUTR3D \cite{chen2023futr3d}, which show the state-of-the-art accuracy in the 3D object detection task. 
% DeepFusion \cite{li2022deepfusion}, TokenFusion \cite{wang2022multimodal}, and FUTR3D \cite{chen2023futr3d} are designed to fuse multi-modal information and achieve state-of-the-art accuracy in the 3D object detection task. 
We extend these methods to be applicable to 3D human body reconstruction \add{with multi-modal multi-view inputs}, and the frameworks are illustrated in \cref{fig:framework}. For DeepFusion and TokenFusion, we employ the parametric reconstruction pipeline by replacing the detection head of the two methods with linear projection to regress SMPL-X \cite{pavlakos2019expressive} parameters. For FUTR3D, we follow its original design and implement the non-parametric pipeline by adding positional embedding of the human body template into the fused features. Subsequently, we employ a Transformer decoder to directly regress joints and vertices.

DeepFusion employs Transformer to integrate local features extracted from multiple modalities. 
This strategy shows the limited capability of leveraging information from other vertices for global interaction due to the absence of global features. Besides, it requires the LiDAR point clouds as the main modality and cannot work with different combinations of modalities, \eg only RGB images. The framework of DeepFusion for 3D body reconstruction is shown in \cref{fig:framework} (b). 
AdaptiveFusion makes a further step to capture the global context by incorporating global features extracted from GIM and it supports arbitrary combinations of input modalities.
% In addition, the adoption of the non-parametric mechanism enables interactions between vertices, joints, local features, and global features, further enhancing the performance of reconstruction. 

For TokenFusion in \cref{fig:framework} (c), it aims to substitute unimportant modality tokens detected by Score Nets with features from other modality streams, and the scores are learned with inter-modal projections among Transformer layers. The design still cannot support arbitrary numbers and orders of modalities. Also, it is ineffective to incorporate the modality streams in adverse environments as undesirable issues like severe sparsity and temporally flicking of point clouds would lead to fetching wrong image features. 
In addition, the adoption of the non-parametric mechanism enables interactions between vertices, joints, local features, and global features, which can further enhance the reconstruction performance of AdaptiveFusion. 

To enable a unified framework for 3D object detection, FUTR3D first encodes features for each modality individually and then employs a query-based Modality-Agnostic Feature Sampler (MAFS) that works in a unified domain to extract features from different modalities as \cref{fig:framework} (d) suggests. Finally, a transformer decoder operates on a set of 3D queries and performs set predictions of objects. Despite its flexibility for input modalities, FUTR3D still needs to train different networks for different modality combinations. Furthermore, it is unable to handle uncalibrated input modalities, limiting its capability to integrate uncalibrated information from other vehicle sources. 
% To demonstrate the superiority of fusing joint/vertex global features and local features of our approach, we employ the parametric reconstruction pipeline by replacing the detection framework of the two methods with linear projection to regress SMPL-X parameters. For DeepFusion, we directly feed all local features from multiple modalities to a generic Transformer-based block to fuse the information, since the method of fusing image features to point cloud features cannot be extended to scenes with more than two modalities. The framework of DeepFusion for 3D body reconstruction is shown in \cref{fig:framework} (b). For the TokenFusion \cite{wang2022multimodal} method, we implement it without any structure altering since it is designed for multimodal input. Specifically, we plug the scoring net among the Transformer layers to dynamically predict the importance of local feature tokens and substitute inferior tokens with their projections from other modalities as \cref{fig:framework} (c) suggests. 

\section{Experiments}
\label{sec:exp}

\subsection{Experimental Settings}

\subsubsection{Datasets} We conduct the experiments on the large-scale mmWave 3D human body dataset mmBody \cite{chen2022mmbody}, which consists of synchronized and calibrated mmWave radar point clouds and RGBD images in various conditions and mesh annotations for humans in the scenes. Following the general setting, we choose 20 sequences in the lab scenes (Lab1 and Lab2) as the training set while 2 sequences for each scene including labs, furnished, rain, smoke, poor lighting, and occlusion\footnote{Only radar and camera1 are occluded and camera2 can still function.} as the test set. As mentioned above, we randomly split the training set into 31 parts to train a single network, and each part consists of a combination of inputs. For testing, we test the network with all combinations on the entire test set. In addition, we evaluate AdaptiveFusion on the other multi-modal and/or multi-view human datasets, Human3.6M \cite{ionescu2013human3}, HuMMan \cite{cai2022humman} and BEHAVE \cite{bhatnagar22behave}, to demonstrate its adaptability. For the Human3.6M dataset, we conduct mix-training using 3D and 2D training data following \cite{lin2021end}. We follow the common setting where subjects S1, S5, S6, S7, and S8 are used in training, and subjects S9 and S11 for testing. We further evaluate our method on the single-view image dataset 3DPW \cite{von2018recovering} to demonstrate its adaptability in the wild.

\subsubsection{Metrics} To evaluate the performance of the reconstruction on the mmBody dataset, we employ commonly used metrics, Mean Per Joint Position Error (MPJPE) and Mean Per Vertex Error (MPVE), which quantify the average Euclidean distance between the prediction and the ground truth for joints/vertices in each frame. For the HuMMan and BEHAVE datasets, we additionally employ Procrustes Analysis MPJPE (PA-MPJPE) to evaluate the alignment accuracy. For the Human3.6M dataset, we report MPJPE and PA-MPJPE following the P2 protocol \cite{kanazawa2018end}.

\subsubsection{Implementation Details}
We implement all the models using Pytorch. For the mmBody, HuMMan, and BEHAVE datasets, we train all the networks on Nvidia GeForce RTX 3090 GPUs for 50 epochs from scratch with an Adam optimizer and an initial learning rate of 0.001. For the Human3.6M dataset, we train our model on 8 NVIDIA V100 GPUs for 200 epochs. For each epoch, our training takes about 2 hours. The loss weights of $\alpha$, $\beta$, and $\gamma$ in our experiments are 1000, 100, and 100, respectively.
% All model weights are randomly initialized, and the image backbone is initialized with ImageNet pre-trained weights. 
% Our AdaptiveFusion applies $L_1$ loss to the reconstructed mesh to constrain the 3D vertices and joints. We also incorporate $L_1$ loss to the 2D joints obtained by projecting predicted 3D joints to the image space using estimated camera parameters. In addition, the coarse meshes of each layer of the Fusion Transformer Module are also supervised by downsampled ground truth meshes using $L_1$ loss to accelerate convergence. 

\subsection{Experimental Results}

\subsubsection{Effectiveness Analysis}
\chen{
We comprehensively analyze the effectiveness of our method. Experimental results demonstrate that AdaptiveFusion can not only adapt to arbitrary modal inputs but also achieve effective fusion by selecting informative features in various environments.
}

\begin{figure}[t]
\centering
    \includegraphics[width=\linewidth]{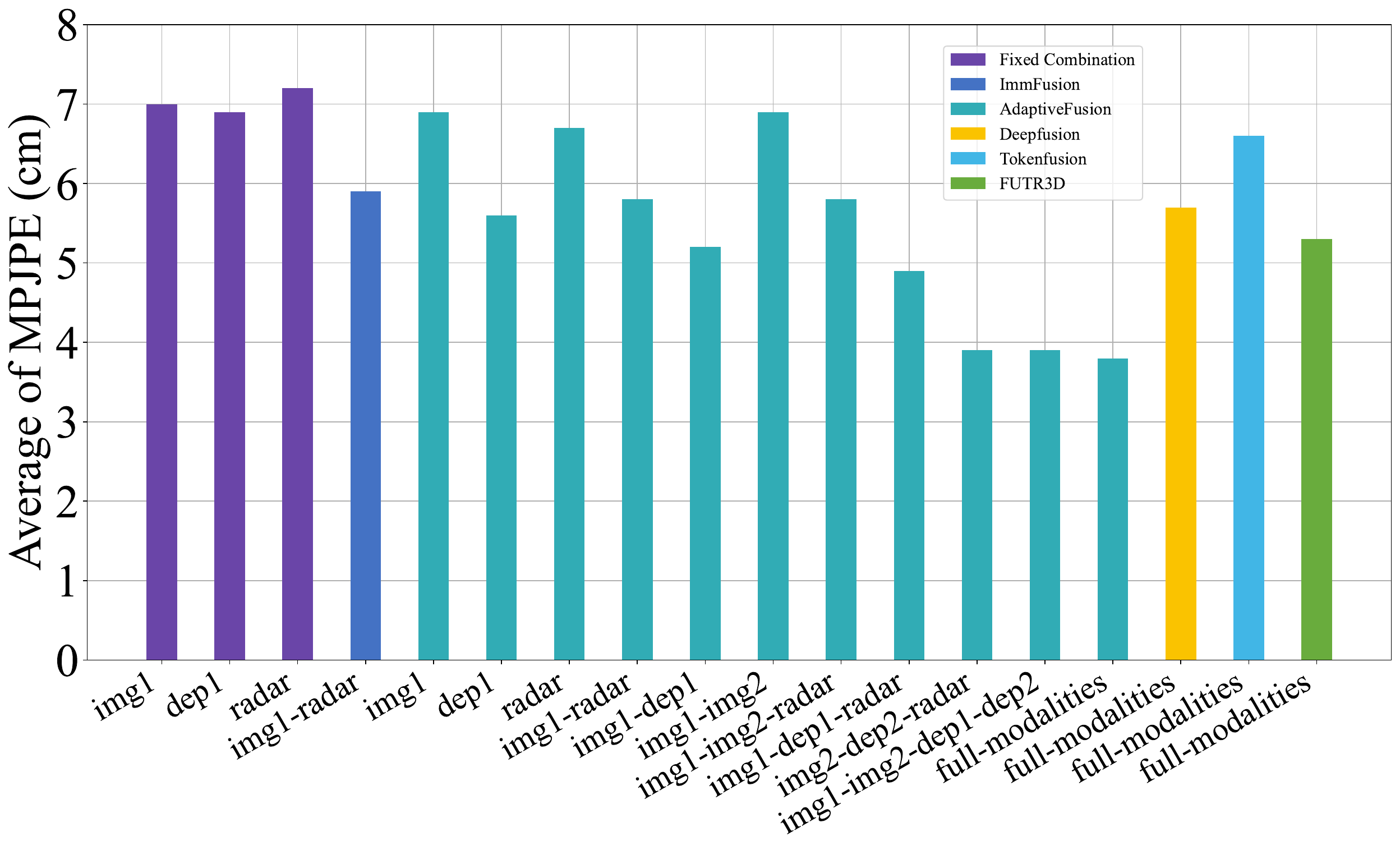}
    \caption{Comparison of AdaptiveFusion using different input combinations with other methods on the mmBody dataset. Img1 and dep1 denote the RGB images and depth point clouds from the first viewpoint. Img2 and dep2 are from the second viewpoint (no adverse conditions for this viewpoint).}
    \label{fig:mpjpe}
\end{figure}

\noindent \textbf{Effectiveness in View and Modality Fusion.}
% Fig2, more modality, better fusion results.
We evaluate AdaptiveFusion using different input combinations from the mmBody dataset as reported in \cref{tab:errors} and average errors are shown in \cref{fig:mpjpe}. We can see that the error decreases with more modalities and views added, which demonstrates our model can integrate the strengths of each modality effectively.

\noindent \textbf{Effectiveness in Arbitrary Input Combinations.}
Our experiments show that AdaptiveFusion can effectively adapt to different input combinations and even improves performance from single-modal approaches.
% Fig2: better than model trained on fixed combination 
We compare its testing results using a single-modal input with models trained only using the fixed single modality. 
As shown in \cref{fig:mpjpe}, AdaptiveFusion trained with different input combinations performs better than models trained with the fixed modality. Notice that AdaptiveFusion only accesses 1/31 of the training data for one combination. This demonstrates the significant advantage of our combination training strategy, which enhances the performance across various modal combinations.
% We can see that AdaptiveFusion exhibits high adaptability in achieving high-quality 3D reconstruction and consistently outperforms fixed-combination methods across different combinations. 

\begin{figure*}[ht]
\centering
    \includegraphics[width=\linewidth]{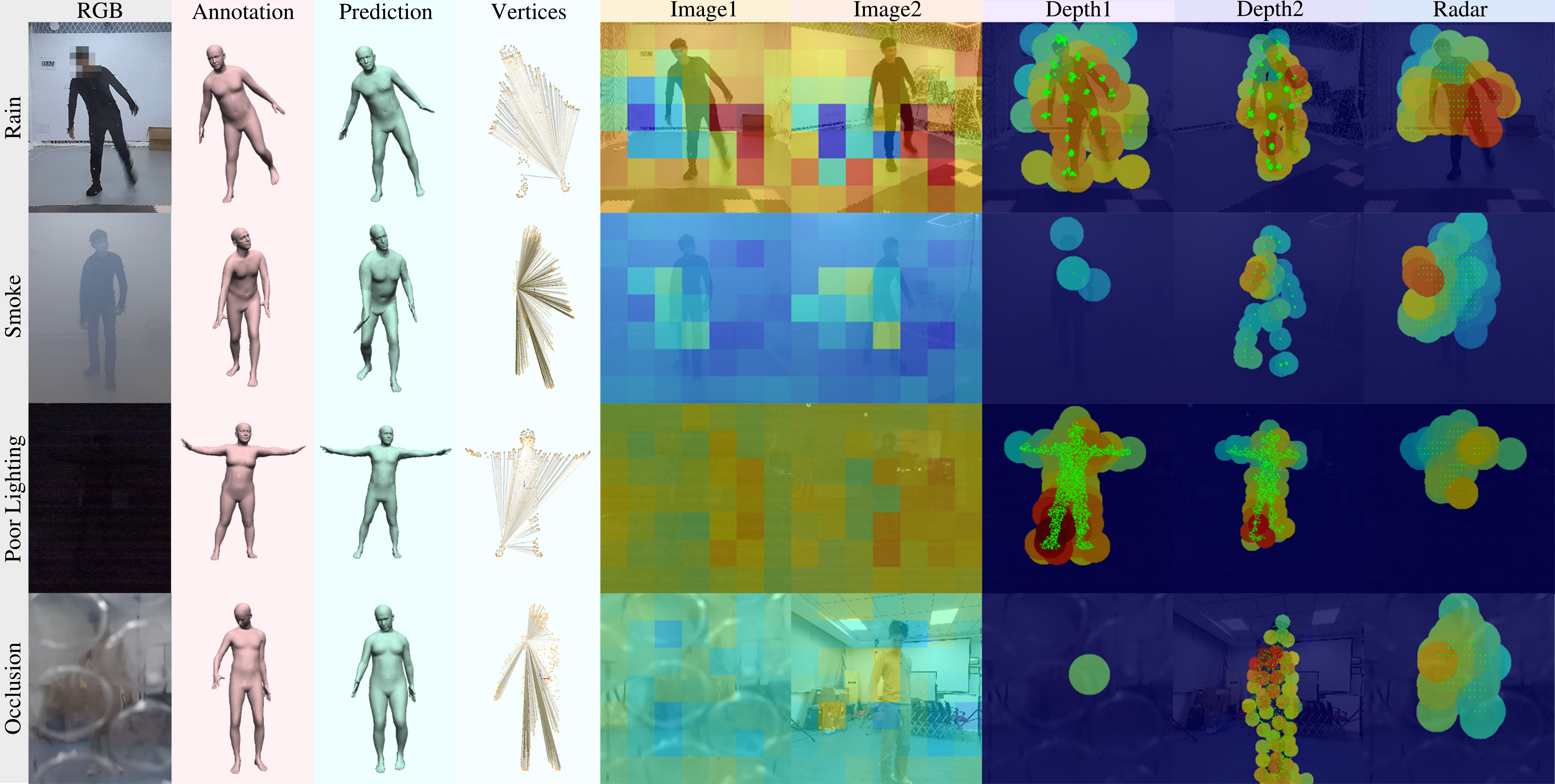}
    \caption{Qualitative results. Each row represents an adverse weather scene (rain, smoke, poor lighting, and occlusion) and each column shows the reconstructed mesh and attention weights, respectively. From top to bottom, weights are for the estimation of the left ankle, right elbow, right ankle, and right shoulder. The darker color in the Vertices column indicates larger attention weights. The reddish color indicates larger attention weights and the bluish color smaller from the Image1 to Radar columns.}
    \label{fig:goodcase}
\end{figure*}

\noindent \textbf{Attention Interactions.}
\cref{fig:goodcase} shows some reconstructed meshes from AdaptiveFusion for different poses and subjects in different scenarios. Overall, the reconstructed meshes for most samples are close to the ground truth. 
We further analyze the attention weights extracted from the last layer of FTM to better understand the effect of AdaptiveFusion in learning interactions between multi-modal multi-view features. \cref{fig:goodcase} shows the visualization of the attention map between a specific joint and other token features including vertices and local features of each modality. For example, at the first row, the depth point clouds are contaminated with rain, leading to the estimation for the left ankle attending more to the image region and the radar point cloud of the left leg. 
In the smoke scene at the second row, both image and depth modalities are severely degraded, which forces the right elbow prediction more reliant on information from the radar point cloud and other vertices. In the poor lighting scene at the third row, the images are highly noisy and the radar points for the right ankle are missing. In this case, the estimation relies more heavily on the depth points of the right leg to accurately regress the location. At the fourth row, camera1 is significantly impaired by the occlusion, and AdaptiveFusion utilizes more information from camera2 and radar to predict the right shoulder position. 
Overall, AdaptiveFusion is able to adapt to various sensor degradation scenarios and effectively integrate relevant information from different modalities to make accurate predictions.

\subsubsection{Comparison Results}
\chen{
\cref{tab:errors}, \cref{tab:human3.6m}, \cref{tab:humman}, \cref{tab:behave}, and \cref{tab:3dpw} summarize the main results of models tested on different datasets. Compared with existing fusion solutions and baselines, our approach performs better in every aspect: in addition to the superior fusion effect of multi-modal inputs by utilizing the complementary feature of each modality, it also achieves significant performance from multi-view inputs. We provide more quantitative comparisons in the supplementary material. 
}

\noindent \textbf{Comparison with Multi-Modal Fusion Methods.}
We compare AdaptiveFusion with the state-of-the-art fusion methods, DeepFusion~\cite{li2022deepfusion}, TokenFusion~\cite{wang2022multimodal} and FUTR3D \cite{chen2023futr3d}. Since the original works are designed for object detection, we extend these approaches to be applicable to 3D human body reconstruction with multi-modal inputs. \cref{tab:errors} includes reconstruction errors with full-modalities input in all scenes of these methods. As we can see, AdaptiveFusion achieves more desirable results in all scenarios. 
DeepFusion exhibits a lower performance compared to AdaptiveFusion in all scenes. This can be mainly attributed to the lack of global features, which reduces global interactions during the fusion stage. 
% shows the limited capability of leveraging information from other vertices for global interaction due to the absence of global features. 
Furthermore, it requires the point clouds as the main modality and cannot work with different combinations of modalities, \eg ~only RGB images.
% Furthermore, despite strengthening interactions between vertices and joints to improve performance, the overall performance of the non-parametric DeepFusion is still inferior to that of AdaptiveFusion. 
TokenFusion aims to substitute unimportant modality tokens detected by Score Nets with features from other modality streams, and the scores are learned with inter-modal projections among Transformer layers. The design still can not support arbitrary numbers and orders of modalities. Besides, it is ineffective to incorporate the modality streams in adverse environments as undesirable issues like severe sparsity and temporally flicking of point clouds would lead to fetching wrong image features. 
% While TokenFusion aims to substitute unimportant token features of each modality with that from other modality streams by using inter-modal projections among Transformer layers. 
% it is ineffective to incorporate the modality streams in adverse environments as undesirable issues like severe sparsity and temporally flicking of point clouds would lead to fetching wrong image features, which ultimately leads to unfavorable results. 
% In terms of TokenFusion, both parametric and non-parametric methods have been observed to demonstrate more limited performance than AdaptiveFusion in adverse environments, particularly in the poor lighting and occlusion scene, where the MPJPE of the two pipelines can reach as high as 9.4cm and 9.7cm. 
For FUTR3D, despite strengthening interactions between vertices and joints to improve performance, the overall performance of the non-parametric FUTR3D is still inferior to that of AdaptiveFusion. 
We also compare AdaptiveFusion with the radar-camera fusion method ImmFusion \cite{chen2023immfusion}. AdaptiveFusion using an image-radar combination achieves comparative or better performance in most scenes. 
Furthermore, AdaptiveFusion with full-modalities input significantly outperforms ImmFusion in all aspects.

\begin{table}[t]
  \centering
  \caption{Results (mm) on the Human3.6M dataset \cite{ionescu2013human3}. * ~denotes pre-trained methods using extra datasets.}
  \resizebox{0.47\textwidth}{!}{
    \begin{tabular}{cccc}
    \toprule
        Input & Methods & MPJPE ↓ & PA-MPJPE ↓ \\
    \midrule
    \multirow{7}[2]{*}{Monocular Image} 
        & SPIN \cite{kolotouros2019learning} & 62.5  & 41.1 \\
        & Pose2Mesh \cite{choi2020pose2mesh} & 64.9  & 46.3 \\
        & Graphormer \cite{lin2021mesh} & 51.2  & 34.5 \\
        & PyMAF \cite{zhang2021pymaf} & 57.7  & 38.7 \\
        % & CLIFF \cite{li2022cliff} & 47.1  & 32.7 \\
        & POTTER \cite{zheng2023potter} & 56.5  & 35.1 \\
        & Kim et al. \cite{kim2023sampling} & 48.3  & 32.9 \\
        & AdaptiveFusion (Ours) & 53.3 & 35.6 \\
    \midrule
    \multirow{7}[2]{*}{Multi-View Images} 
    %     & Deep Triangulation \cite{iskakov2019learnable} * \dag & \textbf{20.8}  & - \\
    % \cmidrule{2-4} 
        & Liang et al. \cite{liang2019shape} & 79.9  & 45.1 \\
        & Shin et al. \cite{shin2020multi} & 46.9  & 32.5 \\
        & Li et al. \cite{li20213d} & 64.8  & 43.8 \\
        & ProHMR \cite{kolotouros2021probabilistic} & 62.2  & 34.5 \\
        & Yu et al. \cite{yu2022multiview} & -     & 33.0 \\
        & Calib-free PaFF \cite{jia2023paff} * & \underline{44.8}  & \textbf{28.2} \\
        & AdaptiveFusion (Ours) & \textbf{43.8} & \underline{31.1} \\
    \bottomrule
    \end{tabular}%
    }
  \label{tab:human3.6m}%
\end{table}%

\noindent \textbf{Comparison with Multi-View Fusion Methods.}
We validate AdaptiveFusion with multi-view images from four views on the Human3.6M dataset as summarized in \cref{tab:human3.6m}. For single-view methods, AdaptiveFusion performs close to the state-of-the-art methods in MPJPE. However, our method with multi-view images significantly enhances performance by fusing information from multiple perspectives. Compared with other multi-view methods, AdaptiveFusion achieves better or on-par performance with a simple framework and without the need for calibration and pre-training. 

\begin{table}[t]
  \centering
  \caption{Results (mm) on the HuMMan  \cite{cai2022humman} datasets.}
  \resizebox{0.47\textwidth}{!}{
    \begin{tabular}{ccccc}
    \toprule
    Methods & Inputs & MPJPE ↓ &  MPVE ↓ &  PA-MPJPE ↓ \\
    \midrule
    HMR \cite{kanazawa2018end} & Monocular Image & 54.78 & -     & 36.14 \\
    VoteHMR \cite{liu2021votehmr} & Monocular Depth & 144.99 & -     & 106.32 \\
    % AdaptiveFusion (Ours) & Monocular Image & 37.47 & 42.6  & 21.56 \\
    AdaptiveFusion (Ours) & Monocular Image & \underline{37.47} & \underline{42.60} & \underline{21.56} \\
    AdaptiveFusion (Ours) & Multi-View Images & \textbf{34.43} & \textbf{42.08} & \textbf{20.82} \\
    \bottomrule
    \end{tabular}}
  \label{tab:humman}%
\end{table}%

\begin{table}[t]
  \centering
  \caption{Results (mm) on the BEHAVE \cite{bhatnagar22behave} datasets.}
  \resizebox{0.47\textwidth}{!}{
    \begin{tabular}{ccccc}
    \toprule
    Methods & Inputs & MPJPE ↓ &  MPVE ↓ &  PA-MPJPE ↓ \\
    \midrule
    Graphormer \cite{lin2021mesh} & Monocular Image & 65.35 & 83.81 & 39.23 \\
    VoteHMR \cite{liu2021votehmr} & Monocular Depth & 63.34 & 72.28 & 40.33 \\
    MV-SMPLify \cite{li20213d} & Multi-View Images & 47.32 & 56.27 & 38.93 \\
    AdaptiveFusion (Ours) & Monocular Image & 64.64 & 82.14 & 40.98 \\
    AdaptiveFusion (Ours) & Monocular Depth & 45.12 & 60.15 & 35.35 \\
    AdaptiveFusion (Ours) & Multi-View Images & \underline{44.16} & \underline{51.36} & \underline{27.33} \\
    AdaptiveFusion (Ours) & Multi-View Multi-Modal & \textbf{31.08} & \textbf{37.89} & \textbf{20.82} \\
    \bottomrule
    \end{tabular}}
  \label{tab:behave}%
\end{table}%

\begin{table}[t]
  \centering
  \caption{Results (mm) on the 3DPW dataset \cite{von2018recovering}.}
    \begin{tabular}{c|cc}
    \toprule
    Methods & MPJPE ↓ &  PA-MPJPE ↓ \\
    \midrule
    SPIN \cite{kolotouros2019learning} & 96.9  & 59.2 \\
    Pose2Mesh \cite{choi2020pose2mesh} & 89.5  & 56.3 \\
    Graphormer \cite{lin2021mesh} & 74.7  & 45.6 \\
    PyMAF \cite{zhang2021pymaf} & 92.8  & 58.9 \\
    SMPLer-X \cite{cai2024smpler} & 75.2  & 50.5 \\
    OSX \cite{lin2023one} & 74.7  & 45.1 \\
    POTTER \cite{zheng2023potter} & 75.0  & \textbf{44.8} \\
    Kim et al. \cite{kim2023sampling} & \textbf{73.9}  & \underline{44.9} \\
    AdaptiveFusion (Ours) & \underline{74.6}  & 45.6 \\
    \bottomrule
    \end{tabular}%
  \label{tab:3dpw}%
\end{table}%

We further validate AdaptiveFusion on the HuMMan and BEHAVE datasets, and the results are summarized in \cref{tab:humman} and \cref{tab:behave}. As the HuMMan dataset does not provide point data currently, we only utilize the image modalities from four viewpoints. 
We compare AdaptiveFusion with other single-modality methods HMR \cite{kanazawa2018end} and VoteHMR \cite{liu2021votehmr} on this dataset\footnote{Results of existing works on the HuMMan dataset are taken from \cite{cai2022humman}.}. HMR is a classic image-based 3D human body reconstruction method and VoteHMR is a recent work that takes point clouds as the input for human mesh recovery. AdaptiveFusion with image modality inputs achieves lower MPJPE and PA-MPJPE compared to the two methods. 

The BEHAVE dataset is a large-scale human-object interactions dataset with challenges of object occlusions and variations in background environments. We evaluate the state-of-the-art monocular-image method Graphormer \cite{lin2021mesh}, depth-based reconstruction method VoteHMR \cite{liu2021votehmr}, and the multi-view image-based method MV-SMPLify \cite{li20213d} on the BEHAVE dataset. 
% Because of these challenges, AdaptiveFusion with image1-only input exhibits relatively high errors. Using multi-view or depth input can significantly improve the performance. 
AdaptiveFusion with single-modality and multi-view inputs exhibits relatively high performance. Furthermore, by combining both multi-view and multi-modal information, AdaptiveFusion with full-modalities input achieves the best results as expected. 
% By combining both multi-view and multi-modal information, AdaptiveFusion shows significant improvements from these works. 

\noindent \textbf{Results in the Wild.}
We further validate AdaptiveFusion with in-the-wild moving monocular images on the 3DPW dataset as shown in \cref{tab:3dpw}. Our method achieves comparative results, demonstrating its capability to work under outdoor mobile perspectives.
It should be noted that our method is designed for multi-view and multi-modal inputs. The result tested with monocular image input is a degraded case for our multi-view model. However, it is comparable with state-of-the-art methods elaborated for single-view images. 

\begin{table}[H]
  \centering
    \caption{\modi{Computational consumption for different input modalities.}}
    \resizebox{0.48\textwidth}{!}{
    \begin{tabular}{c|ccccc}
    \toprule
    Input & Model Params & Num Tokens & Mem Usage & Time (ms) & FPS \\
    \midrule
    img1  & 211.8M & 726 & 3104M & 83.1  & 12.1 \\
    dep1  & 87.1M & 726 & 3114M & 34.6  & 28.9 \\
    radar & 87.1M & 726 & 3114M & 34.6  & 28.9 \\
    img1-radar & 215.1 & 775 & 3116M & 84.4  & 11.8 \\
    img1-dep1-radar & 228.3M & 824 & 3132M & 85.5  & 11.7 \\
    full-modalities & 228.3M & 922 & 3134M & 136.1 & 7.3 \\
    % full-modalities w/ 4M & 228.3M & 726 & 3134M & 134.3 & 7.4 & 3.5 \\
    % full-modalities w/ ResNet & 118.1M & 922 & 2794M & 72.7 & 13.7 & 3.4 \\
    \bottomrule
    \end{tabular}%
    }
  \label{tab:time}%
\end{table}%

\noindent \textbf{Time Consumption.}
We provide results of time consumption \modi{of our method with different inputs} in \cref{tab:time}. Our model consumes affordable computational resources and can achieve real-time performance. 
\modi{With the increasing number of modalities and input data size, we can 1) design an algorithm to identify key image views based on their importance and enable the network to selectively process a limited number of views to control the computational load; 2) utilize the Mamba \cite{gu2023mamba} structure, a method based on the State Space Model with linear complexity of $O(n)$, to reduce memory consumption compared to Transformer architectures, which have quadratic complexity of $O(n^2)$. 3) employ the deformable attention mechanism \cite{zhu2020deformable} to adaptively adjust the attention area, effectively reducing memory usage.}

% \begin{table}[t]
%   \centering
%   \caption{Ablation study on the mmBody dataset.}
%   \resizebox{0.35\textwidth}{!}{
%     \begin{tabular}{c|cc}
%     \toprule
%     Methods & MPJPE ↓ &  MPVE ↓ \\
%     \midrule
%     AdaptiveFusion-w/o-LF & 6.2   & 8.0 \\
%     AdaptiveFusion-w/o-GIM & 4.7   & 6.1 \\
%     AdaptiveFusion-w/o-Masking & 5.3   & 7.1 \\
%     AdaptiveFusion & \textbf{3.8}   & \textbf{4.9} \\
%     \midrule
%     \end{tabular}%
%   }
%   \label{tab:ablation}%
% \end{table}%

\begin{table*}[t]
  \centering
  \caption{Ablation study on the mmBody dataset.}
  \resizebox{\textwidth}{!}{
    \begin{tabular}{c|cc|cc|cc|cc|cc|cc|cc|cc}
    \toprule
    \multicolumn{1}{c|}{\multirow{2}[2]{*}{Methods}} & \multicolumn{6}{c|}{Basic Scenes}             & \multicolumn{8}{c|}{Adverse Environments}                     & \multicolumn{2}{c}{\multirow{2}[2]{*}{Average}} \\
          & \multicolumn{2}{c}{\textbf{Lab1}} & \multicolumn{2}{c}{\textbf{Lab2}} & \multicolumn{2}{c|}{\textbf{Furnished}} & \multicolumn{2}{c}{\textbf{Rain}} & \multicolumn{2}{c}{\textbf{Smoke}} & \multicolumn{2}{c}{\textbf{Poor Lighting}} & \multicolumn{2}{c|}{\textbf{Occlusion}} & \multicolumn{2}{c}{} \\
    \midrule
    AdaptiveFusion-w/o-LF & 4.6   & 5.9   & 4.3   & 4.8   & 5.1   & 6.3   & 5.3   & 6.6   & 7.9   & 10.4  & 6.6   & 8.2   & 9.8   & 14.0  & 6.2   & 8.0 \\
    AdaptiveFusion-w/o-GIM & 3.4   & 4.6   & 3.7   & 4.8   & 4.1   & 5.2   & 4.1   & 5.3   & 6.2   & 7.9   & 3.9   & 4.7   & 7.1   & 10.1  & 4.7   & 6.1 \\
    AdaptiveFusion-w/o-MMM & 3.5   & 4.7   & 3.6   & 4.7   & 4.2   & 5.4   & 3.9   & 5.1   & 6.6   & 8.8   & 5.2   & 6.6   & 10.3  & 14.7  & 5.3   & 7.1 \\
    AdaptiveFusion & \textbf{3.1} & \textbf{4.1} & \textbf{3.3} & \textbf{4.2} & \textbf{3.7} & \textbf{4.4} & \textbf{3.4} & \textbf{4.4} & \textbf{5.2} & \textbf{6.4} & \textbf{3.5} & \textbf{4.3} & \textbf{4.5} & \textbf{6.6} & \textbf{3.8} & \textbf{4.9} \\
    \bottomrule
    \end{tabular}%
    }
  \label{tab:ablation}%
\end{table*}%

\subsection{Ablation Study}
In this section, we conduct a comprehensive study as reported in \cref{tab:ablation} to validate the effectiveness of the local features, Global Integrated Module (GIM), and modality masking module of AdaptiveFusion. 

\subsubsection{Effectiveness of Local Features} 
% The local features, which directly affect the quality and details, play a very important role in reconstruction tasks. 
To analyze the effectiveness of the local features, we compared the results of the original AdaptiveFusion with its variation AdaptiveFusion-w/o-LF, in which the cluster features and grid features are removed from the backward computation graph. As indicated in \cref{tab:ablation}, the mean errors of AdaptiveFusion-w/o-LF are obviously greater than AdaptiveFusion. 
Despite the assistance of MMM, the errors in extreme conditions like occlusion are even worse than those of the Radar-Only method. These results strongly support our motivation of utilizing local features to benefit the quality of reconstruction.
% which strongly supports our motivation of utilizing local features to benefit the quality and details of reconstruction.

\subsubsection{Effectiveness of Global Integrated Module} 
In our framework, GIM serves as a mixer to integrate global features of multi-modal input. Instead of element-wise addition or channel-wise concatenation, GIM uses learnable parameters to control the weights of global features from different modalities. 
In extreme scenarios in which some sensors may not function at all, such as the smoke and occlusion scenes where camera1 is completely occluded, 
GIM allows the model to select the most informative features from the global features of other modalities, resulting in improved accuracy and robustness of reconstruction.
AdaptiveFusion-w/o-GIM is on par with AdaptiveFusion in the basic scenes where the proportion of valid information from RGB, depth, and mmWave sensors is balanced. Simultaneously, with MMM, AdaptiveFusion-w/o-GIM performs well in smoke and poor lighting scenes as well. However, in the extreme scenarios in which some sensors may not function at all, such as the occlusion scene where camera1 is completely occluded, GIM allows the model to select the most informative features from the global features of other modalities, resulting in improved accuracy and robustness of 3D human body reconstruction.

\subsubsection{Effectiveness of Modality Masking Module} 
As stated above, the training set only consisting of basic scene data would force the Transformer module to pay more attention to a single modality, which leads to a rapid decline in performance when the sensor fails in adverse environments. 
% Clearly demonstrated in \cref{tab:ablation}, the masking module eliminates the bias of training data and significantly improves performance in adverse environments. 
As shown in \cref{tab:ablation}, AdaptiveFusion-w/o-MMM can achieve similar errors with AdaptiveFusion in the basic scenes. However, in adverse environments, AdaptiveFusion outperforms AdaptiveFusion-w/o-MMM significantly, demonstrating the importance of MMM in improving the performance of reconstruction. In particular, MMM gains over more than 50\% reconstruction error reduction, and the mean joint/vertex errors for AdaptiveFusion can reach as low as 3.8cm/4.9cm. This is mainly attributed to the fact that MMM forces the Transformer module to lean more attention on the other camera and radar to select helpful features.

\begin{figure}[t]
\centering
    \includegraphics[width=\linewidth]{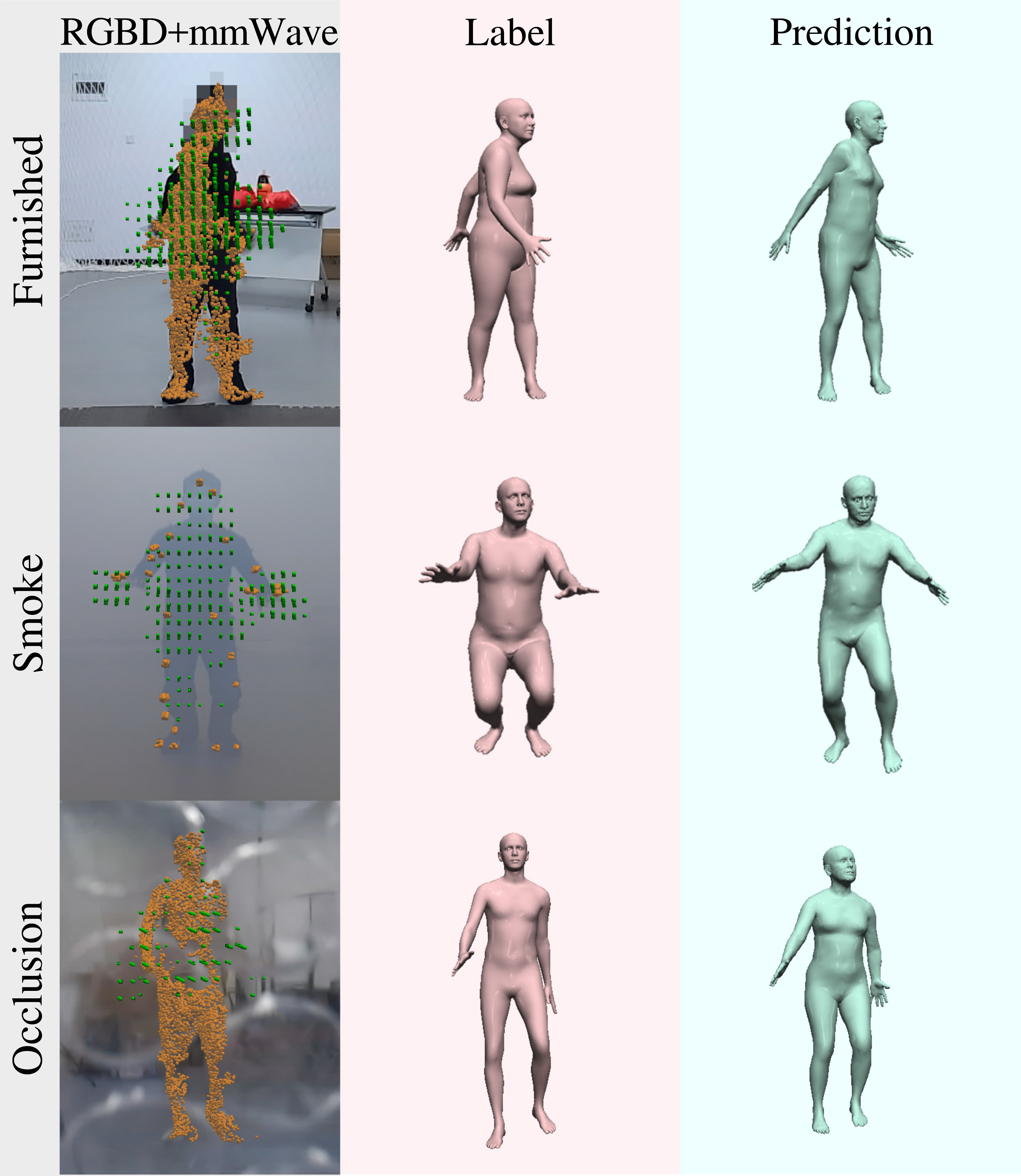}
    \caption{Failure cases in the furnished, smoke, and occlusion scene. (Radar points are in green and depth points are in orange. Depth points in the occlusion scene are from the other viewpoint which is not occluded.)}
    \label{fig:badcase}
\end{figure}

\begin{table}[t]
  \centering
  \caption{Ablation study on the masking strategy.}
    \begin{tabular}{cccc}
    \toprule
    Methods & Mask Proportion & MPJPE ↓ & MPVE ↓ \\
    \midrule
    AdaptiveFusion-w/-4M & / & 4.2   & 5.5 \\
    AdaptiveFusion-w/-MMM & 10\% & 4.1   & 5.3 \\
    AdaptiveFusion-w/-MMM & 30\% & \textbf{3.8} & \textbf{4.9} \\
    AdaptiveFusion-w/-MMM & 50\% & 4.0   & 5.2 \\
    \bottomrule
    \end{tabular}%
  \label{tab:mask}%
\end{table}%

\subsubsection{Ablation on Masking Strategy}
\modi{We provide more ablation results for the masking strategy as shown in \cref{tab:mask}. We adopt 4M \cite{mizrahi20244m}, which decodes only a randomly selected subset of tokens to optimize the training efficiency. 
% Following the 4M approach, we sample the number of input tokens for each modality using a symmetric Dirichlet distribution with concentration parameter $\alpha$. Once the per-modality token counts are determined, tokens from different modalities are sampled uniformly at random. This token selection mechanism ensures an efficient training process while preserving the ability of the model to integrate and learn from multi-modal information. 
Despite the improved efficiency, it exhibits lower accuracy than AdaptiveFusion. This discrepancy can be attributed to the reduced number of tokens, which potentially limits the capacity of the model to capture sufficient information for accurate learning.}
For the mask proportion, we set it to 30\% in the experiments of our main paper as it achieves the best accuracy.

\subsection{Limitations}
Though with the masking module, our model has gained a certain level of generalization ability, it still exhibits relatively high errors in furnished, smoke, and occlusion conditions on the mmBody dataset due to the sensor defects and data imbalance as exampled in \cref{fig:badcase}. In the furnished and smoke scenario, the poor fusion performance can be attributed to the background noise of images and defects of sensors like missing point clouds and multi-path effects, respectively. 
In the occlusion scenario, the subject is much taller than other subjects in the training set \add{and do not wear black clothing like other volunteers when collecting data, leading to the reconstructed shape appears shorter than the ground truth. This issue can be addressed by expanding the diversity of the data. Contrastive and predictive multi-modal pre-training are also promising solutions.
Additionally, constrained by the data limitation, we could only conduct experiments of multi-modal multi-view fusion in the indoor scenes. We leave the extension of our method to outdoor environments as future work.}

% While in the occlusion scenario, where only camera1 is completely occluded, AdaptiveFusion is expected to utilize information from camera2 for better reconstruction. However, the results in this scene are not as satisfactory compared with that in basic scenes. \cref{fig:badcase} (c) illustrates the failure case of reconstruction in the occlusion scene. Despite the fact that the point cloud from depth2 covered the feet, the reconstructed result was shorter compared with the ground truth; and the reconstruction error sharply increases when using only image2 data as input. We attribute this to the fact that the volunteers in the occlusion scene are taller than those in the training set and do not wear black clothing like other volunteers when collecting data, leading to poor reconstruction results. This issue can be addressed by expanding the diversity of the training data. 

% Additionally, the fusion methodology employed by AdaptiveFusion relies on the calibration and synchronization of different sensors, which can result in failure in scenarios where the alignment parameters are unknown, such as dynamic multi-view camera scenes. Incorporating physical constraints of the human body to constrain multi-view reconstruction results may potentially settle this problem. 

\section{Conclusions}
\label{sec:con}

In this paper, we introduce AdaptiveFusion, a generic adaptive 3D reconstruction framework that is capable of processing \modi{arbitrary combinations of given sensor inputs} to achieve robust and precise reconstruction performance. Our approach uses a Transformer network to fuse both global and local features, enabling AdaptiveFusion to adaptively handle arbitrary sensor inputs and accommodate noisy modalities. Furthermore, AdaptiveFusion demonstrates strong adaptability, achieving competitive results even with limited training data for each input combination. We investigate various fusion approaches and demonstrate that AdaptiveFusion outperforms multi-view, LiDAR-camera, and other state-of-the-art multi-modal fusion methods in various environments. 
% Nevertheless, In the future, we plan to improve our ImmFusion approach to push the border of reconstruction from mmWave signals and overcome shortages of current fusion models, like \placeholder{xxx}.

% Though with the masking module our model has gained a certain level of generalization ability, it does not achieve zero-shot capability: it exhibits relatively high error in furnished, smoke, and occlusion conditions on the mmBody dataset due to the sensor defects and data imbalance. Contrastive and predictive multi-modal pre-training are promising solutions.

\bibliographystyle{IEEEtran}
\bibliography{ref}
% \input{sections/biography}

% \vfill

\end{document}